\pdfoutput=1

\documentclass[11pt]{article}

\usepackage[]{acl}

\usepackage{times}
\usepackage{latexsym}

\usepackage[T1]{fontenc}

\usepackage[utf8]{inputenc}

\usepackage{microtype}
\usepackage{enumitem}
\usepackage{amstext}
\usepackage{booktabs}
\usepackage{graphicx}
\usepackage{mathtools}
\usepackage{algorithmic}
\usepackage{algorithm}
\usepackage{amsfonts,amssymb}
\usepackage{multirow}
\usepackage{caption}
\usepackage{subcaption}
\usepackage{array}
\usepackage{bm}

\usepackage{CJKutf8}

\title{PeerDA: Data Augmentation via Modeling Peer Relation for Span Identification Tasks\thanks{~~This work was supported by Alibaba Group through Alibaba Research Intern Program. It was also partially supported by a grant from the Research Grant Council of the Hong Kong Special Administrative Region, China (Project Code: 14200620). $^\dagger$ This work was done when Weiwen Xu was an intern at Alibaba DAMO Academy. $^\ddagger$ Xin Li is the corresponding author.}}

\author{
Weiwen Xu\raisebox{4pt}{\small $12$,}$^\dagger$ \quad Xin Li\raisebox{4pt}{\small $2$,}$^\ddagger$ \quad Yang Deng\raisebox{4pt}{\small $1$} \quad \textbf{Wai Lam}\raisebox{4pt}{\small $1$} \quad \textbf{Lidong Bing}\raisebox{4pt}{\small $2$}\\
\raisebox{4pt}{\small $1$}The Chinese University of Hong Kong \\
\raisebox{4pt}{\small $2$}DAMO Academy, Alibaba Group \\
{\tt \{wwxu,wlam\}@se.cuhk.edu.hk} \quad {\tt ydeng@nus.edu.sg}\\
{\tt \{xinting.lx,l.bing\}@alibaba-inc.com}
}

\date{}

\begin{document}
\maketitle

\begin{abstract}
Span identification aims at identifying specific text spans from text input and classifying them into pre-defined categories. 
Different from previous works that merely leverage the Subordinate (\textsc{Sub}) relation (i.e. \textit{if a span is an instance of a certain category}) to train models, this paper for the first time explores the Peer (\textsc{Pr}) relation, which indicates that \textit{two spans are instances of the same category and share similar features}. Specifically, a novel \textbf{Peer} \textbf{D}ata \textbf{A}ugmentation (PeerDA) approach is proposed which employs span pairs with the \textsc{Pr} relation as the augmentation data for training.
PeerDA has two unique advantages:
  (1) There are a large number of \textsc{Pr} span pairs for augmenting the training data.
  (2) The augmented data can prevent the trained model from over-fitting the superficial span-category mapping by pushing the model to leverage the span semantics.
  Experimental results on ten datasets over four diverse tasks across seven domains demonstrate the effectiveness of PeerDA.
  Notably, PeerDA achieves state-of-the-art results on six of them.\footnote{Our code and data are available at \url{https://github.com/DAMO-NLP-SG/PeerDA}} 
\end{abstract}


\section{Introduction}
\begin{figure}
    \centering
    \includegraphics[scale=0.19]{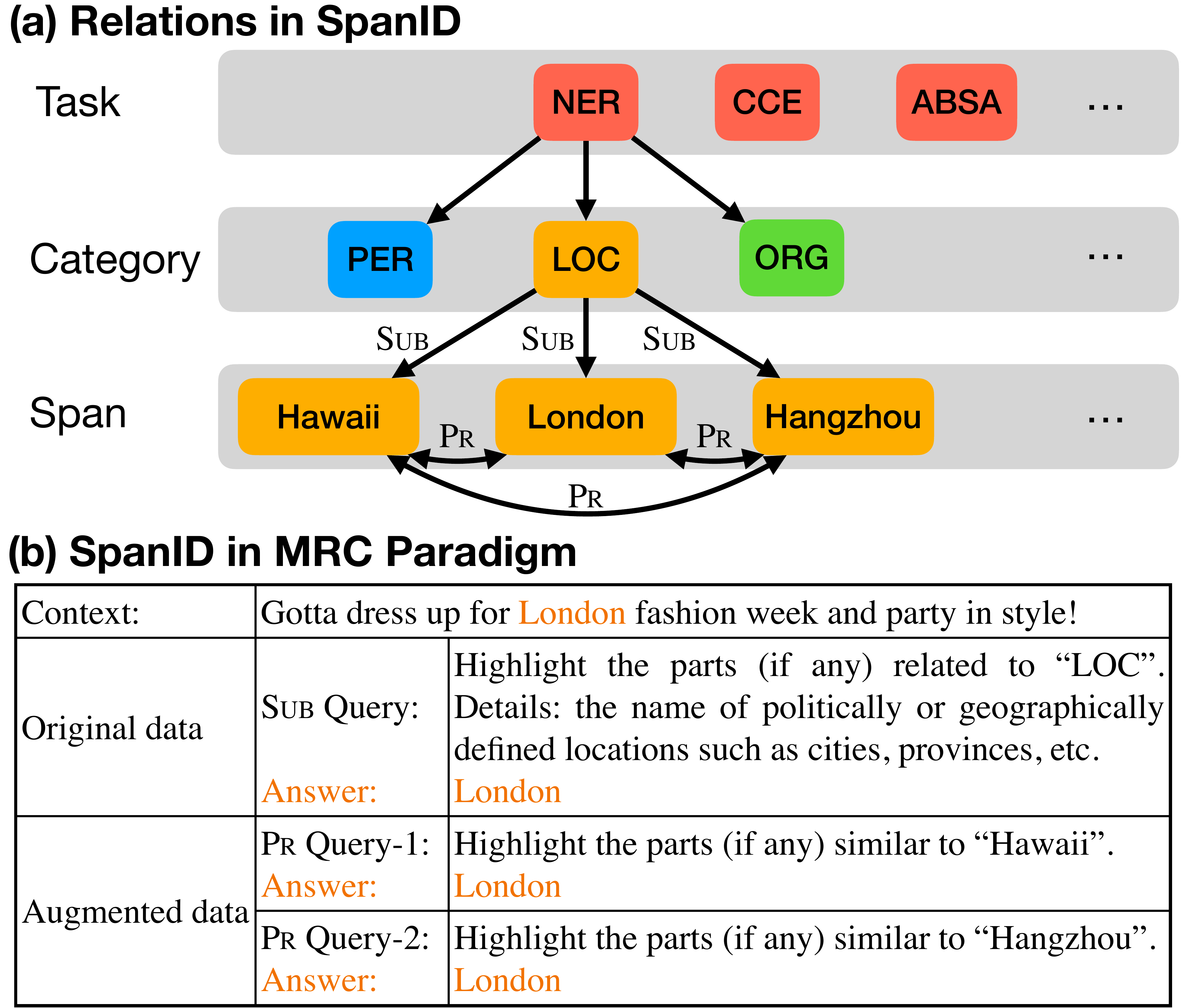}
    \caption{(a) Illustrations of Subordinate (\textsc{Sub}) and Peer (\textsc{Pr}) relations in SpanID tasks. (b) The constructions of augmented data with \textsc{Pr} relations in MRC paradigm. We use NER here for demonstration purposes.}
    \label{fig:example}
    \vspace{-11pt}
\end{figure}

Span Identification (SpanID) is a family of Natural Language Processing (NLP) tasks with the goal of detecting specific spans from the input text and further classifying them into pre-defined categories~\cite{papay-etal-2020-dissecting}.
It serves as the initial step for complex text analysis by narrowing down the search scopes of important spans, which holds a pivotal position in the field of NLP~\cite{ding-etal-2021-nerd,xu-etal-2021-better}.
Recently, different domain-specific SpanID tasks, such as social media Named Entity Recognition (NER) \cite{derczynski-etal-2017-results}, Aspect-Based Sentiment Analysis (ABSA) \cite{liu2012sentiment}, Contract Clause Extraction (CCE) \cite{chalkidis2017extracting}, Span Based Propaganda Detection (SBPD) \cite{da-san-martino-etal-2019-fine} and Argument Extraction \cite{cheng-etal-2020-ape}, have emerged for various NLP applications.
Precisely, as shown in Figure \ref{fig:example} (a), the process of SpanID can be reinterpreted as extracting \textbf{span-category} Subordinate (\textsc{Sub}) relation --- \textit{if a span in the input text is an instance of a certain category}.
Early works \cite{chiu-nichols-2016-named} typically tackle SpanID tasks as a sequence tagging problem, where the \textsc{Sub} relation is recognized via predicting the category for each input token under certain context. 
Recently, to better utilize category semantics, many efforts have been made on reformulating SpanID tasks as a Machine Reading Comprehension (MRC) problem \cite{liu-etal-2020-event,yang-etal-2021-enhanced}.
As shown by the example in Figure \ref{fig:example} (b), such formulation first creates a \textsc{Sub} query for each category and then recognizes the \textsc{Sub} relation by detecting relevant spans in the input text (\textit{i.e.}, context) as answers to the category query.

However, only leveraging the \textsc{Sub} relation in the training data to build SpanID models may suffer from two limitations: 
1) \textbf{Over-fitting}: With only \textsc{Sub} relation, SpanID models tend to capture the superficial span-category correlations. Such correlations may misguide the models to ignore the semantics of the given span but make predictions based on the memorized span-category patterns, which hurts the generalization capability of the models.  
2) \textbf{Data Scarcity}: For low-resource scenarios or long-tailed categories, the number of span-category pairs with \textsc{Sub} relation (\textsc{Sub} pairs) could be very limited and insufficient to learn a reliable SpanID model.


In this paper, we explore the \textbf{span-span} Peer (\textsc{Pr}) relation to alleviate the above limitations. Specifically, the \textsc{Pr} relation indicates that \textit{two spans are two different instances of the same category}. 
The major difference between \textsc{Pr} relation and \textsc{Sub} relation is that the former one intends to correlate two spans without giving the categories they belong to.
For example, in Figure~\ref{fig:example} (a), "Hawaii" and "London" are connected with the \textsc{Pr} relation because they are instances of the same category.
By jointly recognizing \textsc{Sub} relation and \textsc{Pr} relation in the input text, the model is enforced to favor the usage of span semantics instead of span-category patterns for prediction, reducing the risk of over-fitting. 
In addition, the number of span-span pairs with the \textsc{Pr} relation (\textsc{Pr} pairs) grows quadratically over the number of \textsc{Sub} pairs. 
Therefore, we can still construct a reasonable number of training data with \textsc{Pr} pairs for categories having insufficient examples.


In this paper, with the aim of leveraging the \textsc{Pr} relation to enhance SpanID models, we propose a Peer Data Augmentation (\textit{PeerDA}) approach that treats \textsc{Pr} pairs as a kind of augmented training data.
To achieve this, as depicted in Figure~\ref{fig:example} (b), we extend the usage of the original training data into two views.
The first view is the \textsc{Sub}-based training data. It is used to directly solve the SpanID tasks by extracting the \textsc{Sub} relation, which is the typical formulation of MRC-based approaches. 
The second view is the \textsc{Pr}-based training data. It is our augmentation to enrich the semantics of spans by extracting the \textsc{Pr} relation in the original training data, where one span is used to identify its peer from the input context.
Note that our \textsc{Pr}-based training data can be easily formulated into the MRC paradigm. Therefore, the knowledge learned from such augmentation data can be directly transferred to enhance the model's capability to capture \textsc{Sub} relation (\textit{i.e.}, the SpanID tasks).

To better accommodate the MRC-style \textsc{Sub} and \textsc{Pr} data, we develop a stronger and more memory-efficient MRC model. Compared to the designs in~\citet{li-etal-2020-unified}, our model introduces a bilinear component to calculate the span scores  and consistently achieves better performance with a 4 times smaller memory consumption. Besides, we propose a margin-based contrastive learning strategy to additionally model the negative spans to the query (\textit{e.g.}, when querying the context in Figure \ref{fig:example}  for ``ORG'' entities, ``London'' becomes a negative span) so that the spans from different categories are separated more apart in the semantic space.


We evaluate the effectiveness of PeerDA on ten datasets across seven domains, from four different SpanID tasks, namely, NER, ABSA, CCE, and SBPD.
Experimental results show that extracting \textsc{Pr} relation benefits the learning of semantics and encourages models to identify more possible spans.
As a result, PeerDA is a new state-of-the-art (SOTA) method on six SpanID datasets.
Our analyses further demonstrate the capability of PeerDA to alleviate scarcity and over-fitting issues.

Our contributions are summarized as follows:
\begin{itemize}[leftmargin=*,topsep=4pt]
\setlength{\itemsep}{0pt}
\setlength{\parskip}{0pt}
\setlength{\parsep}{0pt}
    \item We propose a novel PeerDA approach to tackle SpanID tasks via augmenting training data with \textsc{Pr} relation.
    \item We conduct extensive experiments on ten datasets, including four different SpanID tasks across seven domains, and achieve SOTA performance on six SpanID datasets.
    \item PeerDA is more effective in low-resource scenarios or long-tailed categories and thus, it alleviates the scarcity issue. Meanwhile, jointly recognizing the \textsc{Sub} and \textsc{Pr} relations makes the MRC model rely less on memorizing the \textsc{Sub} patterns in the training set for inferring the span label, which prevents over-fitting.
\end{itemize}


\section{Related Work}
\paragraph{DA for SpanID:}
DA, which increases the diversity of training data at a low cost, is a widely-adopted solution to address data scarcity \cite{feng-etal-2021-survey}. 
In the scope of SpanID, existing DA approaches aim to introduce more span-category patterns, including:
(1) \textit{Word Replacement} either replaces or paraphrases some context tokens using simple rules~\cite{dai-adel-2020-analysis,xiang2021lexical} and strong language models~\cite{kobayashi-2018-contextual, wu2019conditional,li-etal-2020-conditional,yoo-etal-2021-gpt3mix-leveraging}, or applies synonym dictionaries or masked language models to replace the labeled tokens with other tokens of the same type~\cite{wei-zou-2019-eda,zhou-etal-2022-melm}.
(2) Fine-grained Augmentation Data Generation first trains an auto-regressive language model, and then leverages the model to generate new sentences with entity tags as a special kind of tokens~\cite{ding-etal-2020-daga,liu-etal-2021-mulda}.
(3) \textit{Self-training} is to continually train the model on its predicted data~\cite{xie2019unsupervised,xie2020self,wang2020adaptive,zhou-etal-2023-improving,tan-etal-2023-class}, while 
consistency training also leverages unlabeled data by imposing regularization on the predictions~\cite{zhou-etal-2022-conner}
(4) \textit{Distantly Supervised Training} focuses on leveraging external knowledge to roughly label spans in the target tasks~\cite{bing-etal-2013-wsdm,bing-etal-2015-improving,xu-etal-2023-sampling}. \citet{huang-etal-2021-shot} leverage Wikipedia to create distant labels for NER. \citet{chen-etal-2021-data} transfer data from high-resource to low-resource domains. \citet{jain-etal-2019-entity,li2020unsupervised} tackle cross-lingual NER by projecting labels from high-resource to low-resource languages, which is particularly common in real applications~\cite{kruengkrai-etal-2020-improving}.
Differently, the motivation of PeerDA is to leverage the augmented data to enhance models' capability on semantic understanding by minimizing(maximizing) the distances between semantically similar(distant) spans.

\paragraph{MRC:}
MRC is to extract an answer span from a relevant context conditioned on a given query.
It is initially designed to solve question answering tasks~\cite{hermann2015teaching}, while recent trends have shown great advantages in formulating NLP tasks as MRC problems.
In the context of SpanID, \citet{li-etal-2020-unified,xu2022clozing,xu2022mpmr} address the nested NER issues by decomposing nested entities under multiple queries. \citet{mao2021joint,zhang-etal-2021-aspect} tackle ABSA  in a unified MRC framework. \citet{hendrycks2021cuad} tackle CCE with MRC to deal with the extraction of long clauses.
Moreover, other tasks such as relation extraction \cite{li-etal-2019-entity}, event detection \cite{liu-etal-2020-event,liu-etal-2021-machine}, and summarization \cite{mccann2018natural} are also reported to benefit from the MRC paradigm.

\section{PeerDA}
\paragraph{Overview of SpanID:}
Given the input text $\bm{X}=\{x_1,...,x_n\}$,
SpanID is to detect all appropriate spans $\{\bm{x}_{k}\}_{k=1}^K$ and classify them with proper labels $\{y_k\}_{k=1}^K$, where each span $\bm{x}_{k}=\{x_{s_k}, x_{s_k+1},...,x_{e_k-1},x_{e_k}\}$ is a subsequence of $\bm{X}$ satisfying $s_k \le e_k$ and the label comes from a predefined category set $Y$ (e.g. "Person" in NER).

\subsection{Training Data Construction}
The training data $\mathcal{D}$ consists of two parts:
(1) The \textsc{Sub}-based training data $\mathcal{D^{\textsc{Sub}}}$, where the query is about a category and the MRC context is the input text.
(2) The \textsc{Pr}-based training data $\mathcal{D^{\textsc{Pr}}}$ is constructed with \textsc{Pr} pairs, where one span is used to create the query and the input text containing the second span serves as the MRC context.
\subsubsection{\textsc{Sub}-based Training Data}
First, we need to transform the original training examples into (query, context, answers) triples following the paradigm of MRC \cite{li-etal-2020-unified}.
To extract the \textsc{Sub} relation between categories and relevant spans, a natural language query $\bm{Q}^{\textsc{Sub}}_y$ is constructed to reflect the semantics of each category $y$.
Following \citet{hendrycks2021cuad}, we include both category mention $\tt [Men]_y$ and its definition $\tt [Def]_y$ from the annotation guideline (or Wikipedia if the guideline is not accessible) in the query to introduce more comprehensive semantics:
\begin{equation}
\small
\begin{aligned}
    \bm{Q}^{\textsc{Sub}}_y=\;&\tt{Highlight\;the\;parts\;(if\;any)}\\
    &\tt{related\;to\;[Men]_y.\;Details:[Def]_y}.
\end{aligned}
\end{equation}

Given the input text $\bm{X}$ as the context, the answers to $\bm{Q}^{\textsc{Sub}}_y$ are the spans belonging to category $y$.
Then we can obtain one MRC example denoted as ($\bm{Q}^{\textsc{Sub}}_y$, $\bm{X}$, $\{\bm{x}_{k}\;|\;\bm{x}_{k}\in \bm{X},y_k=y\}_{k=1}^K$).
To guarantee the identification of all possible spans, we create $|Y|$ training examples by querying the input text with each pre-defined category.

\subsubsection{ \textsc{Pr}-based training data}
\label{sec:pr_data}
To construct augmented data that derived from the \textsc{Pr} relation, we first create a category-wise span set $\mathcal{S}_y$ that includes all training spans with category $y$:
\begin{equation}
\small
    \mathcal{S}_y = \{\bm{x}_{k}\; |\; (\bm{x}_{k}, y_k) \in \mathcal{D^{\textsc{Sub}}}, y_k = y \}
\end{equation}
Obviously, any two different spans in $\mathcal{S}_y$ have the same category and shall hold the \textsc{Pr} relation.
Therefore, we pair every two different spans in $\mathcal{S}_y$ to create a peer set $\mathcal{P}_y$:
\begin{equation}
    \mathcal{P}_y = \{(\bm{x}^q,\bm{x}^a)\; |\; \bm{x}^q,\bm{x}^a \in \mathcal{S}_y ,\\ \bm{x}^q \ne \bm{x}^a \}
\end{equation}

For each \textsc{Pr} pair $(\bm{x}^q,\bm{x}^a)$ in $\mathcal{P}_y$, we can construct one training example by constructing the query with the first span $\bm{x}^q$:
\begin{equation}
\small
\begin{aligned}
    \bm{Q}^{\textsc{Pr}}_y=\;&\tt{Highlight\;the\;parts\;(if\;any)\;} \\
    & \tt{similar\;to\;}\bm{\mathit{x^q}}.
\end{aligned}
\end{equation}
Then we treat the text $\bm{X}^a$ containing the second span $\bm{x}^a$ as the MRC context to be queried and $\bm{x}^a$ as the answer to $\bm{Q}^{\textsc{Pr}}_y$. Note that there may exist more than one span in $\bm{X}^a$ satisfying \textsc{Pr} relation with $\bm{x}^q$, we set all of them as the valid answers to $\bm{Q}^{\textsc{Pr}}_y$, yielding one training example
($\bm{Q}^{\textsc{Pr}}_y$, $\bm{X}^a$, $\{\bm{x}^a_{k}\;|\;\bm{x}^a_{k}\in \bm{X}^a,y^a_k=y\}_{k=1}^K$) of our PeerDA.

Theoretically, given the span set $\mathcal{S}_y$, there are only $|\mathcal{S}_y|$ \textsc{Sub} pairs in the training data but we can obtain $|\mathcal{S}_y| \times (|\mathcal{S}_y| - 1)$ \textsc{Pr} pairs to construct $\mathcal{D^{\textsc{Pr}}}$.
Such a large number of augmented data shall hold great potential to enrich spans' semantics. 
However, putting all \textsc{Pr}-based examples into training would exacerbate  the skewed data distribution issue since the long-tailed categories get fewer \textsc{Pr} pairs for augmentation and also increase the training cost.
Therefore, as the first step for DA with the \textsc{Pr} relation, we propose three augmentation strategies to control the size and distribution of augmented data.

\paragraph{PeerDA-Size:}
This is to increase the size of augmented data while keeping the data distribution unchanged.
Specifically, for each category $y$, we randomly sample $\lambda|\mathcal{S}_y|$ \textsc{Pr} pairs from $\mathcal{P}_y$. Then we collect all sampled \textsc{Pr} pairs to construct $\mathcal{D^{\textsc{Pr}}}$, where $\lambda$ is the DA rate to control the size of $\mathcal{D^{\textsc{Pr}}}$.

\paragraph{PeerDA-Categ:}
Categories are not evenly distributed in the training data, and in general SpanID models perform poorly on long-tailed categories.
To tackle this, we propose PeerDA-Categ to augment more training data for long-tailed categories.
Specifically, let $y^*$ denote the category having the largest span set of size $|\mathcal{S}_{y^*}|$. We sample up to $|\mathcal{S}_{y^*}| - |\mathcal{S}_y|$ \textsc{Pr} pairs from $\mathcal{P}_y$ for each category $y$ and construct a category-balanced training set $\mathcal{D^{\textsc{Pr}}}$ using all sampled pairs. 
Except for the extreme cases where $|\mathcal{S}_y|$ is smaller than $\sqrt{|\mathcal{S}_{y^*}|}$, we would get the same size of the training data for each category after the augmentation, which significantly increases the exposure for spans from the long-tailed categories.

 \paragraph{PeerDA-Both (The final version of PeerDA):} 
To take advantage of the above two strategies, we further propose PeerDA-Both to maintain the data distribution while effectively increasing the size of training data. In PeerDA-Both, we randomly sample $\max( \lambda|\mathcal{S}_{y^*}| + (|\mathcal{S}_{y^*}| - |\mathcal{S}_y|),0)$ \textsc{Pr} pairs from $\mathcal{P}_y$ for each category $y$ to construct $\mathcal{D^{\textsc{Pr}}}$, where $\lambda|\mathcal{S}_{y^*}|$ determines the size of the augmented data, and $|\mathcal{S}_{y^*}| - |\mathcal{S}_y|$ controls the data distribution.
\subsubsection{Data Balance}
We combine the $\mathcal{D^{\textsc{Sub}}}$ and the $\mathcal{D^{\textsc{Pr}}}$ created above as the final training data.
Since an input text usually mentions spans from a few categories, when converting the text into the MRC paradigm, many of the $|Y|$ examples are unanswerable. If a SpanID model is trained on this unbalanced data, then the model may favor the majority of the training examples and output an empty span. 
To balance answerable and unanswerable examples, we follow \citet{hendrycks2021cuad} to randomly remove some unanswerable examples from the training data.

\begin{table*}[t]
    \centering
    \small
    \begin{tabular}{@{}l||p{13mm}<{\centering}p{11mm}<{\centering}p{7mm}<{\centering}p{12mm}<{\centering}cccccc@{}}\toprule
        Task  &  \multicolumn{5}{c}{NER} & \multicolumn{2}{c}{ABSA} & \multicolumn{2}{c}{SBPD} & CCE\\ 
                 \cmidrule(lr){1-1}\cmidrule(lr){2-6}\cmidrule(lr){7-8}\cmidrule(lr){9-10}\cmidrule(lr){11-11}
        Dataset & \textbf{OntoNotes5}  & \textbf{WNUT17} & \textbf{Movie} & \textbf{Restaurant} & \textbf{Weibo} & \textbf{Lap14}  & \textbf{Rest14} & \textbf{News20} & \textbf{Social21} & \textbf{CUAD}\\ 
        Domain & \textit{mixed}  & \textit{social} & \textit{movie} & \textit{restaurant} & \textit{social} & \textit{laptop}  & \textit{restaurant} & \textit{news} & \textit{social} & \textit{legal} \\ 
        \# Train    & 60.0k  & 3.4k & 7.8k & 7.7k & 1.3k & 2.7k   & 2.7k    & 0.4k     & 0.7k & 0.5k \\ 
        \# Test     & 8.3k  & 1.3k & 2.0k & 1.5k &  0.3k & 0.8k   & 0.8k    & 75 (\textit{dev}) & 0.2k & 0.1k\\
        \# Category & 11     & 6    & 12   & 8 &  4    & 1 / 3  &  1 / 3    &14        & 20 & 41\\ \bottomrule
    \end{tabular}
    \caption{Statistics on the ten SpanID datasets. Note that 1 / 3 denotes that there is 1 category in ATE and 3 categories in UABSA. \textit{dev} denotes that we evaluate News20 on the dev set.}
    \label{tab:statis}
    \vspace{-10pt}
\end{table*}

\subsection{Model Architecture}
\begin{figure}
    \centering
    \includegraphics[scale=0.3]{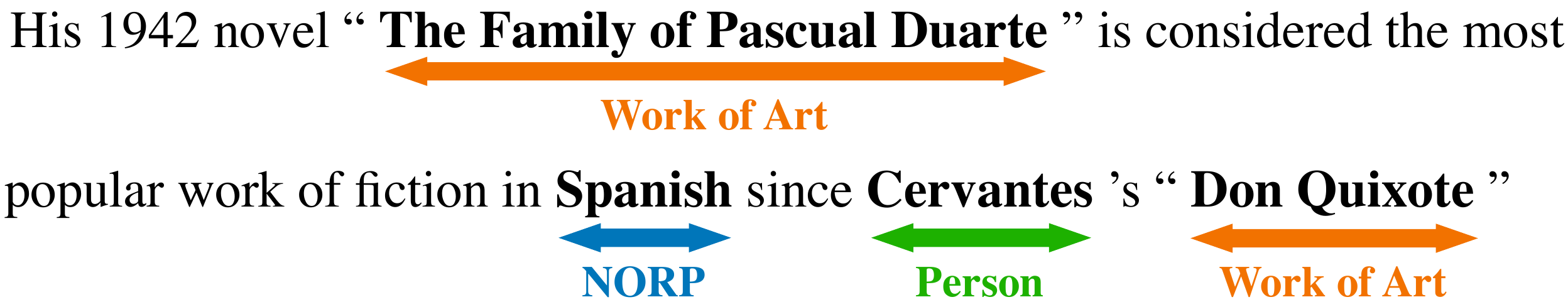}
    \caption{Example of extracting multiple spans in NER.}
    \label{fig:multi}
    \vspace{-10pt}
\end{figure}

As shown in Figure \ref{fig:multi}, to achieve the detection of multiple spans for the given query, we follow \citet{li-etal-2020-unified} to build the MRC model. Compared to the original designs, we further optimize the computation of span scores  following a \textit{general} way of \citet{luong-etal-2015-effective,xu2022clozing}.

Specifically, the base model consists of three components: an encoder, a span predictor, and a start-end selector. First, given the concatenation of the query $\bm{Q}$ and the context $\bm{X}$ as the MRC input $\overline{\bm{X}}=\{{\tt [CLS]},\bm{Q},{\tt [SEP]},\bm{X},{\tt [SEP]}\}$, where ${\tt [CLS]}, {\tt [SEP]}$ are special tokens, the encoder would encode the input text into hidden states $\bm{H}$:
\begin{equation}
\small
\setlength{\abovedisplayskip}{5pt}
\setlength{\belowdisplayskip}{1pt}
    \bm{H} = \textsc{Encoder}(\overline{\bm{X}})
\end{equation}

Second, the span predictor consists of two binary classifiers, one to predict whether each context token is the start index of the answer, and the other to predict whether the token is the end index:
\begin{equation}
\small
\setlength{\abovedisplayskip}{5pt}
\setlength{\belowdisplayskip}{5pt}
    P_\text{start}=\bm{H} W^s \;\;\;\;\; P_\text{end}=\bm{H} W^e 
\end{equation}
where $W^s,  W^e \in \mathbb{R}^{d \times 2}$ are the weights of two classifiers and $d$ is the dimension of hidden states.
The span predictor would output multiple start and end indexes for the given query and context.

Third, the start-end selector matches each start index to each end index and selects the most possible spans from all combinations as the outputs. Different from the \textit{concat} way that would create a large $\mathbb{R}^{|\overline{\bm{X}}| \times |\overline{\bm{X}}| \times 2d}$-shape tensor  \cite{li-etal-2020-unified}, we leverage a \textit{general} way following \citet{luong-etal-2015-effective,xu2022clozing} to compute the span score, consuming fewer resources for better training efficiency:

\begin{equation}
\label{eq:score}
\small
    P_\text{s,e} = \textit{FFN}(\bm{H}_{s})^T\bm{H}_{e}
\end{equation}
where \textit{FFN} is the feed-forward network  \cite{NIPS2017_3f5ee243}, $P_\text{s,e}$ denotes the likelihood of $\overline{\bm{X}}_{s:e}$ to form a possible answer.

\subsection{Training Objective}
The standard objective is to minimize the cross-entropy loss (CE) between above three predictions and their corresponding ground-truth labels, i.e.,  $ Y_\text{start}, Y_\text{end}, Y_\text{s,e}$ \cite{li-etal-2020-unified}:
\begin{equation}
\small
\setlength{\abovedisplayskip}{5pt}
\setlength{\belowdisplayskip}{5pt}
\begin{aligned}
    \mathcal{L}_{mrc}=& ~ \text{CE}(\sigma(P_\text{start}), Y_\text{start}) + \text{CE}(\sigma(P_\text{end}), Y_\text{end}) \\& + \text{CE}(\sigma(P_\text{s,e}), Y_\text{s,e})
    \end{aligned}
\end{equation}
where $\sigma$ is the sigmoid function.

However, these objectives only capture the semantic similarity between the query and positive spans  (i.e., the span instances of the query category). 
In this paper, we propose to explicitly separate the query and its negative spans (i.e., the span instances of other categories) apart with a margin-based contrastive learning strategy, for better distinguishing the spans from different categories.

Specifically, given the MRC input $\overline{\bm{X}}$ with query of category $y$, there may be multiple positive spans $\overline{\mathcal{X}}^+=\{\overline{\bm{x}}_{k} \in \overline{\bm{X}}, y_k=y\}$ and negative spans $\overline{\mathcal{X}}^-=\{\overline{\bm{x}}_{k'} \in \overline{\bm{X}}, y_{k'} \ne y\}$. We leverage the following margin-based contrastive loss to penalize negative spans \cite{chechik2010large}:
\begin{equation}\label{eq:ct}
    \small
    \mathcal{L}_{ct} = \max_{\overline{\bm{x}}_{k} \in \overline{\mathcal{X}}^+ \atop \overline{\bm{x}}_{k'} \in \overline{\mathcal{X}}^-}max(0, M- (\sigma(P_{s_k,e_k}) - \sigma(P_{s_{k'},e_{k'}})))
\end{equation}
where $M$ is the margin term, $max(\cdot,\cdot)$ is to select the larger one from two candidates, and the span score $P_{s_k,e_k}$ can be regarded as the semantic similarity between the query and the target span $\overline{\bm{x}}_k$. Note that our contrastive loss maximizes the similarity difference between the query and the most confusing positive and negative span pairs (\textit{Max-Min}), which we demonstrate to be effective in Sec.~\ref{sec:ablation}. 

Finally, the overall training objective is:
\begin{equation}
    \small
    \label{eqn:loss}
    \mathcal{L} =  \mathcal{L}_{mrc} + \alpha \mathcal{L}_{ct}
\end{equation}
where $\alpha$ is the balance rate.

\section{Experimental Setup}
\subsection{Tasks}
Note that PeerDA is a method for augmenting training data, it is not applied to test sets during evaluation. Therefore, we only use the \textsc{Sub}-based test data to evaluate the models' capability to extract spans according to the category.
We conduct experiments on four SpanID tasks from diverse domains, including NER, ABSA, Contract Clause Extraction (CCE), and Span-Based Propaganda Detection (SBPD). The dataset statistics are summarized in Table~\ref{tab:statis}. The detailed task description can be found in Appendix \ref{sec:task_dis}. 
\paragraph{NER:}
It is to detect named entities (i.e. spans) and classify them into entity types (i.e. categories).
We evaluate five datasets, including four English datasets: \textbf{OntoNotes5}\footnote{In order to conduct robustness experiments in Sec. \ref{sec:robust}, we use the datasets from \citet{lin-etal-2021-rockner} with 11 entity types.} \cite{pradhan-etal-2013-towards}, \textbf{WNUT17} \cite{derczynski-etal-2017-results}, \textbf{Movie} \cite{liu2013query}, and \textbf{Restaurant}~\cite{liu2013asgard} and a Chinese dataset \textbf{Weibo} \cite{peng-dredze-2015-named}.
We use span-level micro-averaged Precision, Recall, and F$_1$ as evaluation metrics.


\begin{table*}[t]
    \centering
    \small
    \begin{tabular}{@{}lp{5mm}<{\centering}p{5mm}<{\centering}p{5mm}<{\centering}|p{5mm}<{\centering}p{5mm}<{\centering}p{5mm}<{\centering}|p{5mm}<{\centering}p{5mm}<{\centering}p{5mm}<{\centering}|p{5mm}<{\centering}p{5mm}<{\centering}p{5mm}<{\centering}|p{5mm}<{\centering}p{5mm}<{\centering}p{5mm}<{\centering}@{}}\toprule
        \multirow{2}{*}{Methods}   &  \multicolumn{3}{c}{ \textbf{OntoNotes5}} & \multicolumn{3}{c}{\textbf{WNUT17}}  & \multicolumn{3}{c}{ \textbf{Movie}} & \multicolumn{3}{c}{\textbf{Restaurant}}&  \multicolumn{3}{c}{\textbf{Weibo}}\\
        \cmidrule(lr){2-4}\cmidrule(lr){5-7}\cmidrule(lr){8-10}\cmidrule(lr){11-13}\cmidrule(lr){14-16}
        & P     & R     & F$_1$    & P     & R     & F$_1$    & P     & R     & F$_1$    & P     & R & F$_1$    & P     & R     & F$_1$    \\ \midrule
        &\multicolumn{3}{c|}{\underline{RB-CRF+RM}}&\multicolumn{3}{c|}{\underline{CL-KL}}&\multicolumn{3}{c|}{\underline{T-NER}}&\multicolumn{3}{c|}{\underline{KaNa}}&\multicolumn{3}{c}{\underline{RoBERTa+BS}}\\
        SOTA    & 92.8     &  92.4    & 92.6   & -     & -     & \textbf{60.5}  & -     & - & 71.2  & 80.9 & 80.0 & 80.4& 70.2     & 75.4     & \textbf{72.7}\\\midrule
              &\multicolumn{15}{c}{\underline{\texttt{Base}}}\\
     Tagging    & 91.0 & 91.8 & 91.4  & 62.1 & 48.2 & 54.3 & 73.0 & 72.8 & 72.9 & 80.6 & 80.7 & 80.7 & 70.8 & 71.0 & 70.9\\
     MRC        & 92.4 & 91.8 & 92.1  & 66.4 & 40.7 & 50.5 & 70.3 & 73.3 & 71.8 & 81.4 & 79.9 & 80.6 & 73.6 & 64.4 & 68.7\\ 
     PeerDA   & 91.9 & 92.6 & {92.4}  & 71.1 & 46.9 & {56.5} & 77.9 & 72.3 & {75.0} & 81.3 & 82.8 & {82.1} & 70.0 & 73.3 & {71.6}\\
  \midrule
     &\multicolumn{15}{c}{\underline{\texttt{Large}}}\\
     Tagging    & 93.0 & 92.3 & 92.6  & 69.4 & 46.2 & 55.4 & 74.2 & 74.0 & 74.1 & 80.9 & 82.0 & 81.4 & 71.4 & 69.2 & 70.3\\
     MRC        & 92.8 & 91.8 & 92.3  & 72.4 & 41.7 & 52.9 & 76.7 & 73.2 & 74.9 & 81.6 & 81.7 & 81.7 & 72.2 & 66.8 & 69.4\\
     PeerDA   & 92.8 & 93.7 & {\textbf{93.3}}  & 70.9 & 48.0 & {57.2} & 78.5 & 73.1 & {\textbf{75.7}} & 81.8 & 82.5 & {\textbf{82.2}} & 73.4 & 71.6 & {72.5}\\
 \bottomrule
    \end{tabular}
    \caption{Performance on NER datasets. The best models are bolded.}
    \label{tab:ner}
    \vspace{-8pt}
\end{table*}

\paragraph{ABSA:} We explore two ABSA sub-tasks: \textbf{Aspect Term Extraction (\textbf{ATE})} to only extract aspect terms, and \textbf{Unified Aspect Based Sentiment Analysis} \textbf{(UABSA)} to jointly identify aspect terms and their sentiment polarities.
We evaluate the two sub-tasks on two datasets, including the laptop domain \textbf{Lap14} and restaurant domain \textbf{Rest14}.
We use micro-averaged F$_1$ as the evaluation metric.

\paragraph{SBPD:} It aims to detect both the text fragment where a persuasion technique is used (i.e. spans) and its technique type (i.e. categories). We use \textbf{News20} and \textbf{Social21} from SemEval shared tasks~\cite{da-san-martino-etal-2020-semeval,dimitrov-etal-2021-semeval}. 
For \textbf{News20}, we report the results on its dev set since the test set is not publicly available.
We use micro-averaged Precision, Recall, and F$_1$ as evaluation metrics.

\paragraph{CCE:} It is a legal task to detect and classify contract clauses (i.e. spans) into relevant clause types (i.e. categories), such as "Governing Law".  We conduct CCE experiments using  \textbf{CUAD}~\cite{hendrycks2021cuad}.
We follow \citet{hendrycks2021cuad} to use Area Under the Precision-Recall Curve (AUPR) and Precision at 80\% Recall (P@0.8R) as the evaluation metrics.

\subsection{Implementations}
Since legal SpanID tasks have a lower tolerance for missing important spans, we do not include start-end selector (i.e.  $\text{CE}(P_\text{s,e}, Y_\text{s,e})$ and  $\alpha \mathcal{L}_{ct}$ in Eq. (\ref{eqn:loss})) in the CCE models but follow ~\citet{hendrycks2021cuad} to output top 20 spans from span predictor for each input example in order to extract spans as much as possible. While for NER, ABSA, and SBPD, we use our optimized architecture and objective.

For a fair comparison with existing works, our models utilize BERT \cite{devlin-etal-2019-bert} as the text encoder for ABSA and RoBERTa \cite{liu2019roberta} for NER, CCE, and SBPD. 
Detailed configurations can be found in Appendix~\ref{sec:impl}.

\subsection{Baselines}
Note that our main contribution is to provide a new perspective to treat the \textsc{Pr} relation as a kind of training data for augmentation.
Therefore, we compare with models built on the same encoder-only PLMs \cite{devlin-etal-2019-bert,liu2019roberta}.
We are not focusing on pushing the SOTA results to new heights though some of the baselines already achieved SOTA performance. 

\paragraph{NER:}
We compare with Tagging \cite{liu2019roberta} and MRC \cite{li-etal-2020-unified} baselines.
We also report the previous best approaches for each dataset, including RB-CRF+RM \cite{lin-etal-2021-rockner}, CL-KL \cite{wang-etal-2021-improving}, T-NER \cite{ushio-camacho-collados-2021-ner}  KaNa \cite{nie2021knowledge}, and RoBERTa+BS \cite{zhu-li-2022-boundary}.
\paragraph{ABSA:}
In addition to the MRC baseline, we also compare with previous approaches on top of BERT.
These are SPAN-BERT \cite{hu-etal-2019-open}, IMN-BERT \cite{he-etal-2019-interactive}, RACL \cite{chen-qian-2020-relation} and Dual-MRC \cite{mao2021joint}.

\paragraph{SBPD:}
For \textbf{News20} we only compare with MRC baseline due to the lack of related work. For \textbf{Social21}, we compare with top three approaches on its leaderboard, namely, Volta \cite{gupta-etal-2021-volta}, HOMADOS \cite{kaczynski-przybyla-2021-homados}, and TeamFPAI \cite{hou-etal-2021-fpai}.

\paragraph{CCE:}
We compare with (1) MRC basline, (2) stronger text encoders, including ALBERT \cite{lan2019albert} and DeBERTa \cite{he2020deberta}, (3) the model continually pretrained on contracts: RoBERTa + CP \cite{hendrycks2021cuad} and (4) the model leveraged the contract structure: ConReader \cite{xu-etal-2022-conreader}.
\section{Results}
\subsection{Comparison Results}
\label{sec:result}
\paragraph{NER:}
Table~\ref{tab:ner} shows the performance on five NER datasets. 
Our PeerDA significantly outperforms the Tagging and MRC baselines.
Precisely, compared to RoBERTa$_{\tt base}$ MRC, PeerDA obtains 0.3, 6.0, 3.2, 1.5, and 2.9 F$_1$ gains on five datasets respectively.
When implemented on  RoBERTa$_{\tt large}$, our PeerDA can further boost the performance and establishes new SOTA on three datasets, namely, \textbf{OntoNotes5}, \textbf{Movie}, and \textbf{Restaurant}.
Note that the major improvement of PeerDA over MRC comes from higher Recall. 
It implies that PeerDA encourages models to give more span predictions.
\begin{table}[t]
    \centering
    \small
    \begin{tabular}{@{}lp{11mm}<{\centering}p{7mm}<{\centering}|p{11mm}<{\centering}p{6mm}<{\centering}}\toprule
      \multirow{2}{*}{Methods}  &   \multicolumn{2}{c}{ \textbf{Lap14}}  &   \multicolumn{2}{c}{ \textbf{Rest14}} \\
      \cmidrule(lr){2-3}\cmidrule(lr){4-5}
                            & UABSA & ATE     & UABSA & ATE  \\ \midrule
        SPAN-BERT           & 61.3 & 82.3   & 73.7 & 86.7  \\
        IMN-BERT            & 61.7 & 77.6   & 70.7 & 84.1  \\
        RACL                & 63.4 & 81.8   & 75.4 & 86.4 \\
        Dual-MRC            & 65.9 & 82.5   & \textbf{76.0} & 86.6  \\ \midrule
        MRC (\underline{\texttt{Large}}) & 63.2 & 83.9 & 72.9 & 86.8\\
        PeerDA        & {\textbf{65.9}} & {\textbf{84.6}} & 73.9 & {\textbf{86.8}}\\\bottomrule
    \end{tabular}
    \caption{Performance on two ABSA subtasks on two datasets. Results are averages F$_1$ over 5 runs.}
    \label{tab:ABSA}
    \vspace{-8pt}
\end{table}


\paragraph{ABSA:}
Table~\ref{tab:ABSA} depicts the results on ABSA. 
Compared to previous approaches, PeerDA mostly achieves better results on two subtasks, where it outperforms vanilla MRC by 2.7 and 1.0 F$_1$ on UABSA for two domains respectively.

\begin{table}[]
    \centering
    \small
    \begin{tabular}{@{}l@{\;\;\;}p{4.5mm}<{\centering}p{4.5mm}<{\centering}p{5mm}<{\centering}|p{4.5mm}<{\centering}p{4.5mm}<{\centering}p{5mm}<{\centering}}\toprule
      \multirow{2}{*}{Methods}   & \multicolumn{3}{c}{ \textbf{News20}} & \multicolumn{3}{c}{ \textbf{Social21}} \\
      \cmidrule(lr){2-4}\cmidrule(lr){5-7}
                    & P     & R     & F$_1$    & P    & R    & F$_1$\\ \midrule
         Volta          & -     & -     & -     & 50.1 & 46.4 & 48.2 \\ 
         HOMADOS        & -     & -     & -     & 41.2 & 40.3 & 40.7 \\
         TeamFPAI       & -     & -     & -     & 65.2 & 28.6 & 39.7 \\ \midrule
         MRC (\underline{\texttt{Base}}) & 10.5  & 53.5  & 17.6  & 55.8 & 43.5 & 48.9 \\
         PeerDA        & 21.8 & 31.5  & {\textbf{25.8}}   & 49.4 & 70.6 & {\textbf{58.1}} \\ \bottomrule
    \end{tabular}
    \caption{PeerDA performance on two SBPD datasets.}
    \label{tab:SBPD}
    \vspace{-8pt}
\end{table}

\begin{table}[t]
    \centering
    \small
    \begin{tabular}{@{}p{26mm}ccc}\toprule
    Methods                                 & \#Params  & AUPR  & P@0.8R\\ \midrule
    $\text{ALBERT}_{\tt xxlarge}$           & 223M  & 38.4  & 31.0  \\
    $\text{RoBERTa}_{\tt base}\text{ + CP}$ & 125M  & 45.2  & 34.1  \\
    $\text{RoBERTa}_{\tt large}$            & 355M  & 48.2  & 38.1  \\
    $\text{DeBERTa}_{\tt xlarge}$           & 900M  & 47.8  & 44.0  \\ 
    ConReader$_{\tt large}$                 & 355M  & 49.1  & 44.2 \\\midrule
    MRC (\underline{\texttt{Base}})         & 125M  & 43.6  & 32.2  \\
    PeerDA                                 & 125M  & {\textbf{52.3}}  & \textbf{45.5}  \\ \bottomrule
    \end{tabular}
    \caption{PeerDA performance on CCE.}
    \label{tab:CCE}
    \vspace{-5pt}
\end{table}

\paragraph{SBPD:}
The results of two SBPD tasks are presented in Table~\ref{tab:SBPD}.
PeerDA outperforms MRC by 8.2 and 9.2 F$_1$ and achieves SOTA performance on \textbf{News20} and \textbf{Social21} respectively. 

\paragraph{CCE:}
The results of CCE are shown in Table~\ref{tab:CCE}.
PeerDA surpasses MRC by 8.7 AUPR and 13.3 P@0.8R and even surpasses the previous best model  of larger size ($\text{ConReader}_{\tt large}$) by 3.2 AUPR, reaching SOTA performance on \textbf{CUAD}.

\begin{table}[t]
    \centering
    \small
    \setlength{\tabcolsep}{1.2mm}{
    \begin{tabular}{@{}l|ccccc@{}}
    \toprule
    \textbf{Ablation Type}          & \textbf{NER}  & \textbf{UABSA}& \textbf{SBPD} & \textbf{CCE} &\textbf{Avg.} \\ \midrule
    MRC                             & 72.7          & 68.1          & 33.3          & 43.6          & 54.4  \\
    PeerDA-Size                     & 74.6          & 69.7          & 38.5          & 48.7          & 57.9  \\
    PeerDA-Categ                    & 74.2          & 69.3          & 40.4          & 51.3          & 58.8  \\
    PeerDA-Both (\textbf{final})    & \textbf{75.5}          & \textbf{69.9}          & \textbf{42.0}          & \textbf{52.3}          & \textbf{59.9} \\\bottomrule
    \end{tabular}}
    \caption{Ablation study on data augmentation strategies. The results (F$_1$ for NER, UABSA, and SBPD. AUPR for CCE) are averaged of all datasets in each task.}
    \label{tab:data_abl}
\end{table}

\begin{table}[t]
    \centering
    \small
    \setlength{\tabcolsep}{1.2mm}{
    \begin{tabular}{@{}l|c|cccc@{}}
    \toprule
    \textbf{Ablation Type}  & \textbf{|GPU|}   & \textbf{NER}  & \textbf{UABSA}& \textbf{SBPD} & \textbf{Avg.} \\
    \midrule \multicolumn{6}{c}{ \em \small Calculation of $P_{s,e}$} \\\midrule
    \textit{concat}           & 1x                    & 74.5          & 69.2          & 40.3          &  61.3 \\
    \textit{general} (\textbf{final})& \textbf{0.23x}                 & \textbf{75.0}          & \textbf{69.4}          & \textbf{40.8}      &  \textbf{61.7} \\
    \midrule \multicolumn{6}{c}{ \em \small Contrastive Loss} \\\midrule
    \textit{Average}                 & 0.23x                 & 75.1          & 69.6          & 37.6          &  60.8 \\
    \textit{Max-Min} (\textbf{final})& 0.23x                 & \textbf{75.5}          & \textbf{69.9}          & \textbf{42.0}   & \textbf{62.4}\\\bottomrule
    \end{tabular}}
    \caption{Ablation study on model designs. The F$_1$ scores are averaged of all datasets in each task. The \textbf{|GPU|} column denotes the GPU memory footprint of each variant under the same experimental setup.}
    \label{tab:model_abl}
\end{table}

\subsection{Analysis on Augmentation Strategies}
To explore how the size and category distribution of the augmented data affect the SpanID tasks, we conduct ablation study on the three augmentation strategies mentioned in Sec. \ref{sec:pr_data}, depicted in Table \ref{tab:data_abl}.
Overall, all of the PeerDA variants are clearly superior to the MRC baseline and the PeerDA-both considering both data size and distribution issues performs the best.
Another interesting finding is that PeerDA-Categ significantly outperforms PeerDA-Size on SBPD and CCE. We attribute the phenomenon to the fact that SBPD and CCE have a larger number of categories and consequently, the MRC model is more prone to the issue of skewed data distribution.
Under this circumstance, PeerDA-Categ, the variant designed for compensating the long-tailed categories, can bring larger performance gains over MRC model.
On the other hand, if the skewed data distribution is not severe (e.g. NER), or the category shows a weak correlation with the spans (i.e. UABSA), PeerDA-Size is more appropriate than PeerDA-Categ.

\begin{figure*}
    \centering
    \includegraphics[scale=0.22]{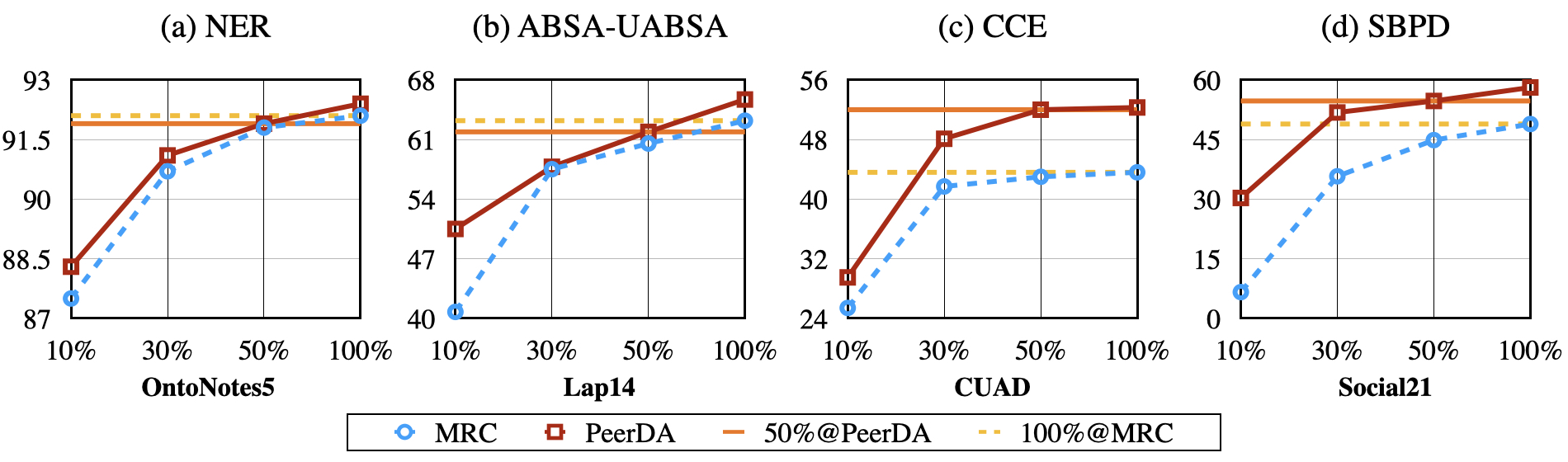}
    \caption{Performance on low-resource scenarios. We select one dataset for each SpanID task and report the test results (AUPR for CCE and F$_1$ for others) from the models trained on different proportions of the training data.} 
    \label{fig:low-resource}
    \vspace{-8pt}
\end{figure*}

\subsection{Analysis on Model Designs}
\label{sec:ablation}
\paragraph{Calculation of $P_{s,e}$} 
(Top part of Table \ref{tab:model_abl}) Under the same experimental setup ($\text{RoBERTa}_{\tt base}$, batch size=32, sequence length=192, fp16), using our \textit{general} method (Eq. (\ref{eq:score})) to compute span score $P_{s,e}$ greatly reduces the memory footprint by more than 4 times with no performance drop, compared to the original \textit{concat} method. Therefore, our \textit{general} method allows a larger batch size for accelerating the training. 

\paragraph{Contrastive Loss} (Bottom part of Table \ref{tab:model_abl})
After we have settled on the \textit{general} scoring function, we further investigate different methods to compute contrastive loss. We find that the \textit{Average} method, which averages similarity differences between the query and all pairs of positive and negative spans, would affect SpanID performance when the task has more long-tailed categories (i.e. SBPD).
While our \textit{Max-Min} (strategy in Eq.(\ref{eq:ct})) is a relaxed regularization, which empirically is more suitable for SpanID tasks and consistently performs better than the \textit{Average} method.

\begin{table}[t]
    \centering
    \small
    \setlength{\tabcolsep}{1mm}{
    \begin{tabular}{@{}lc|c|c|c@{}}\toprule
    \multirow{2}{*}{ SRC $\rightarrow$ TGT }     & \multicolumn{2}{c|}{ $\text{RoBERTa}_{\tt base}$} & \multicolumn{2}{c}{ $\text{RoBERTa}_{\tt large}$}  \\
    \cmidrule(lr){2-3}\cmidrule(lr){4-5}
     & MRC & PeerDA & MRC & PeerDA \\ \midrule

     \textbf{Onto.} $\rightarrow$ \textbf{WNUT17} & 43.1 & \textbf{46.8} & 44.2  & \textbf{46.9} \\
     \textbf{Onto.} $\rightarrow$ \textbf{Rest.}   & 1.6 & \textbf{5.0} & 2.7 & \textbf{11.0} \\
     \textbf{Onto.} $\rightarrow$  \textbf{Movie} & 25.0  & \textbf{26.7}   & 26.7  & \textbf{27.8} \\
    Average   & 23.3   & \textbf{26.2}  & 24.5 & \textbf{28.6}\\
 \bottomrule
    \end{tabular}}
    \caption{F$_1$ scores on NER cross-domain transfer, where models trained on source-domain training data (SRC) are evaluated on target-domain test sets (TGT). }
    \label{tab:NER-cross}
    \vspace{-8pt}
\end{table}

\begin{figure}
    \centering
    \includegraphics[scale=0.5]{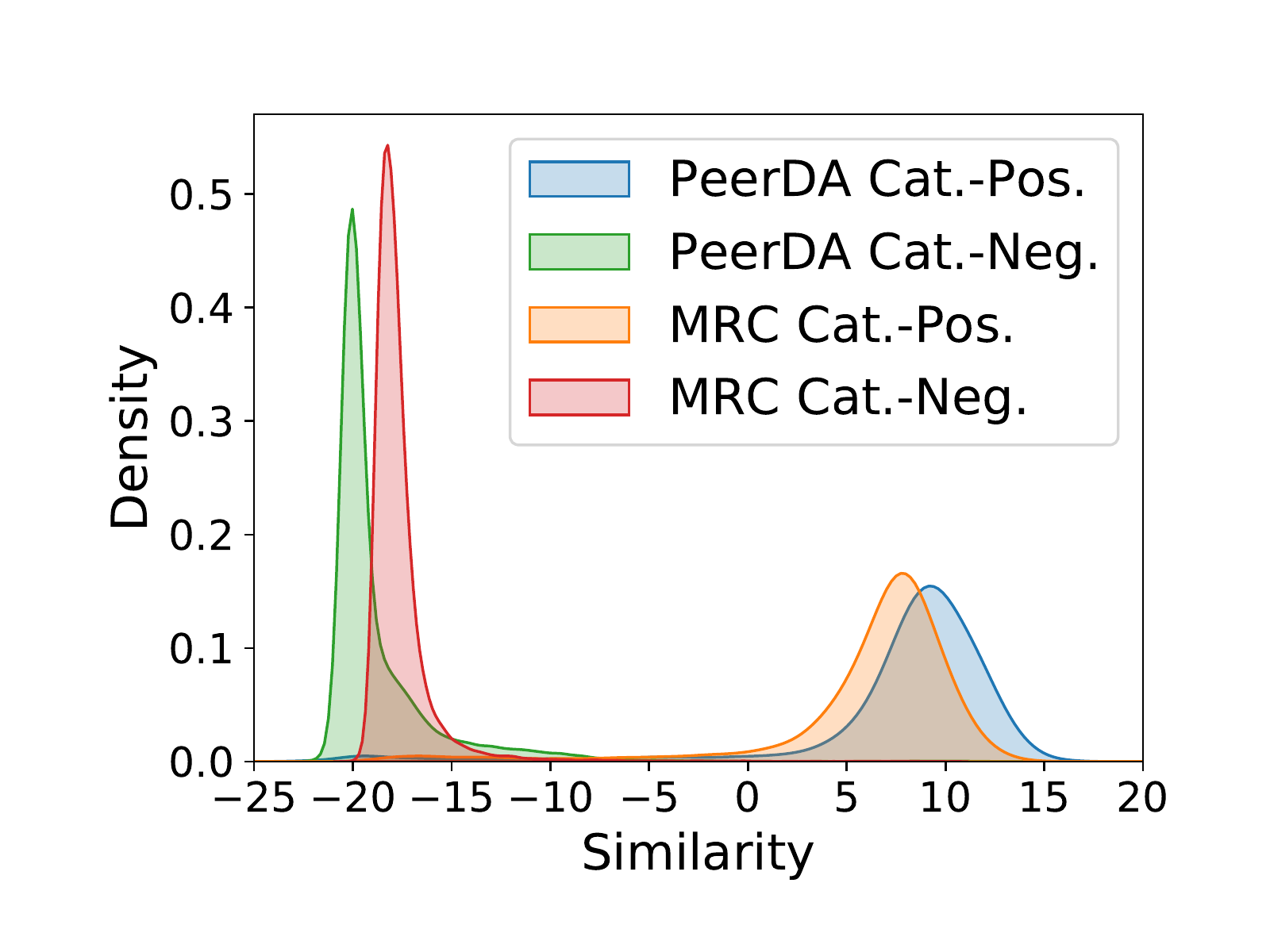}
    \caption{The distribution of similarity score between categories and their corresponding positive/negative spans on \textbf{Ontonotes5} test set.} 
    \label{fig:semantics}
    \vspace{-8pt}
\end{figure}

\section{Further Discussions}
In this section, we make further discussions to bring valuable insights of our PeerDA approach. 

\paragraph{Out-of-domain Evaluation:}
We conduct out-of-domain evaluation on four English NER datasets, where the model is trained on \textbf{OntoNotes5}, the largest dataset among them, and evaluated on the test part of another three datasets.
Since these four datasets are from different domains and differ substantially in their categories, this setting largely eliminates the impact of superficial span-category patterns and thus it can faithfully reflect how well the MRC model exploits span semantics for prediction.
The results are presented in Table~\ref{tab:NER-cross}.
PeerDA can significantly exceed MRC on all three transfer pairs.
On average, PeerDA achieves 2.9 and 4.1 F$_1$ gains over base-size MRC and large-size MRC respectively.
These results verify our postulation that modeling the \textsc{Pr} relation allows models to weigh more on the semantics for making predictions, and thus mitigates the over-fitting issue.

\begin{table*}[]
\fontsize{8.5}{10}\selectfont
    \centering
    \setlength{\tabcolsep}{1mm}{
    \begin{tabular}{c|p{116mm}}
    \toprule
       \multirow{4}{*}{Multiple Labels}         &  \textit{I'm in Atlanta.} \\
          &  \textbf{Gold}: ("Atlanta", {\tt GPE})\\
              &  \textbf{PeerDA}: ("Atlanta", {\tt GPE}); \textcolor{blue}{("Atlanta", {\tt LOC}) (41\%)}\\
              &  \textbf{MRC}:  ("Atlanta", {\tt GPE}) \textcolor{red}{("Atlanta", {\tt LOC})  (3\%)} \\
       \midrule
       \multirow{4}{*}{Incorrect Label}     &  \textit{Why did it take us to get Sixty Minutes to do basic reporting to verify facts?}\\
                       &  \textbf{Gold}: ("Sixty Minutes", {\tt ORG})\\
                        &  \textbf{PeerDA}: \textcolor{blue}{("Sixty Minutes", {\tt WORK\_OF\_ART}) (37\%)} \\
                        &  \textbf{MRC}: \textcolor{red}{("Sixty Minutes", {\tt WORK\_OF\_ART}) (20\%)}\\
       \midrule
       \multirow{4}{*}{Missing Prediction}   & \textit{Coming to a retailer near you, PlayStation pandemonium.} \\
                      &  \textbf{Gold}: ("PlayStation", {\tt PRODUCT})\\
                            &  \textbf{PeerDA}: \textcolor{blue}{$\emptyset$ (19\%)} \\
                            &  \textbf{MRC}: \textcolor{red}{$\emptyset$ (74\%)} \\
        \midrule
       \multirow{4}{*}{Other Errors}     & \textit{I was guarded uh by the British Royal Marines actually because unfortunately they've had now um uh roadside bombs down there not suicide bombs.} \\
                            &  \textbf{Gold}: ("the British Royal Marines",{\tt ORG}) \\
                                    &  \textbf{PeerDA}:  \textcolor{blue}{("Royal Marines",{\tt ORG}) (3\%)}   \\
                                    &  \textbf{MRC}: \textcolor{red}{("Royal Marines",{\tt ORG}) (3\%)} \\

       \bottomrule
    \end{tabular}}
    \caption{Error analysis of base-sized PeerDA and MRC models on \textbf{Ontonotes5} test set. We randomly select 100 examples from the test set and compare the predictions and error percentages of the two models.}
    \label{tab:error}
\end{table*}

\paragraph{Semantic Distance:}
To gain a deeper understanding of the way in which PeerDA enhances model performance, we consider the span score (Eq. \ref{eq:score}) as a measure of semantic similarity between a query and a span. In this context, we can create queries for all categories and visualize the similarity distribution between the categories and their corresponding positive and negative spans on \textbf{Ontonote5} test set.
As shown in Figure \ref{fig:semantics}, 
we can observe that the use of PeerDA leads to an increased semantic similarity between spans and their corresponding categories, resulting in higher confidence in the prediction of correct spans. Furthermore, PeerDA has been shown to also create a larger similarity gap between positive and negative spans, facilitating their distinction.

\paragraph{Low-resource Evaluation:}
We simulate low-resource scenarios by randomly selecting 10\%, 30\%, 50\%, and 100\% of the training data for training SpanID models and show the comparison results between PeerDA and MRC on four SpanID tasks in Figure~\ref{fig:low-resource}.
As can be seen, our PeerDA further enhances the MRC model in all sizes of training data and the overall trends are consistent across the above four tasks.
When training PeerDA with 50\% of the training data, it can reach or even exceed the performance of MRC trained on the full training set.
These results demonstrate the effectiveness of our PeerDA in low-resource scenarios.

\paragraph{Error Analysis:}
In order to know the typical failure of PeerDA, we randomly sample 100 error cases from \textbf{Ontonotes5} test set for analysis.
As shown in Table \ref{tab:error}, there are four major groups:
\begin{itemize}[leftmargin=3mm]
\setlength{\itemsep}{0pt}
\setlength{\parskip}{0pt}
\setlength{\parsep}{0pt}
    \item \textit{Multiple Labels}: PeerDA would assign multiple labels to the same detected span. And in most cases (35/41), this error occurs among similar categories, such as {\tt LOC}, {\tt GPE}, and {\tt ORG}.
    \item \textit{Incorrect Label}: Although spans are correctly detected, PeerDA assigns them the wrong categories. Note that MRC even cannot detect many of those spans (23/37). As a result, PeerDA significantly improves the model's capability to detect spans, but still faces challenges in category classification.
    \item \textit{Missing Prediction}: Compared to MRC, PeerDA tends to predict more spans. Therefore it alleviates the missing prediction issue that MRC mostly suffers.
    \item \textit{Other Errors}: There are several other errors, such as the incorrect span boundary caused by articles or nested entities. 
\end{itemize}

\section{Conclusions}
In this paper, we propose a novel PeerDA approach for SpanID tasks to augment training data from the perspective of capturing the \textsc{Pr} relation.
PeerDA has two unique advantages: (1) It is capable to leverage abundant but previously unused \textsc{Pr} relation as additional training data. (2) It alleviates the over-fitting issue of MRC models by pushing the models to weigh more on semantics. 
We conduct extensive experiments to verify the effectiveness of PeerDA.
Further in-depth analyses demonstrate that the improvement of PeerDA comes from a better semantic understanding capability.
\section*{Limitations}
In this section, we discuss the limitations of this work as follows:
\begin{itemize}[leftmargin=3mm]
\setlength{\itemsep}{0pt}
\setlength{\parskip}{0pt}
\setlength{\parsep}{0pt}
    \item PeerDA leverages labeled spans in the existing training set to conduct data augmentation. This means that PeerDA improves the semantics learning of existing labeled spans, but is ineffective to classify other spans outside the training set.
    Therefore, it would be beneficial to engage outer source knowledge (e.g. Wikipedia), where a variety of important entities and text spans can also be better learned with our PeerDA approach.
    \item Since PeerDA is designed in the MRC formulation on top of the encoder-only Pre-trained Language Models (PLMs) \cite{devlin-etal-2019-bert, liu2019roberta}, it is not comparable with other methods built on encoder-decoder PLMs \cite{yan-etal-2021-unified-generative,chen-etal-2022-lightner,zhang-etal-2021-towards-generative,yan-etal-2021-unified}. It would be of great value to try PeerDA on encoder-decoder PLMs such as BART \cite{lewis-etal-2020-bart} and T5 \cite{raffel2020exploring}, to see whether PeerDA is a general approach regardless of model architecture.
    \item As shown in Table \ref{tab:error}, although PeerDA can significantly alleviate the \textit{Missing Predictions}, the most prevailing error in the MRC model, PeerDA also introduces some new errors, i.e. \textit{Multiple labels} and \textit{Incorrect Label}.
    It should be noted that those problematic spans are usually observed in different span sets, where they would learn different category semantics from their peers.
    Therefore, we speculate that those spans tend to leverage the learned category semantics more than their context information to determine their categories. We hope such finding can shed light on future research to further improve PeerDA.
\end{itemize}

\bibliography{custom}
\bibliographystyle{acl_natbib}
\clearpage
\begin{table*}[]
    \centering
    \small
    \begin{tabular}{@{}l||cccccccccc@{}}\toprule
        Dataset & OnteNote5 & WNUT17 & Movie & Restaurant & Weibo & Lap14  & Rest14 & CUAD & News20 & Social21\\ \midrule
        Query Length & 32       & 32    & 64    & 64 & 64   & 24    & 24    & 256   & 80    & 80\\ 
        Input Length & 160      & 160   & 160   & 128 & 192  & 128   & 128   & 512   & 200   & 200\\
        Batch Size   & 32       & 32    & 32    & 32 & 8   & 16    & 16    & 16    & 16    & 16\\
        Learning Rate & 2e-5    & 1e-5  & 1e-5  & 1e-5 & 1e-5 & 2e-5  & 2e-5  & 5e-5  & 2e-5  & 3e-5\\ 
        $\lambda$     & 1       & 1     & 1     & 1  & 1   & 1     & 1     & -0.5    & 0.5   & 1\\    \bottomrule
    \end{tabular}
    \caption{Hyper-parameters settings.}
    \label{tab:params}
\end{table*}
\appendix
\section{Appendix}
\label{sec:appendix}
\subsection{Task Overview}
\label{sec:task_dis}
We conduct experiments on four SpanID tasks with diverse domains, including Named Entity Recognition (NER), Aspect Based Sentiment Analysis (ABSA), Contract Clause Extraction (CCE) and Span Based Propaganda Detection (SBPD), to show the overall effectiveness of our PeerDA. The dataset statistics are summarized in Table~\ref{tab:statis}.
\paragraph{NER:} It is a traditional SpanID task, where spans denote the named entities in the input text and category labels denote their associated entity types. We evaluate five datasets from four domains:

\begin{itemize}[leftmargin=3mm]
    \item \textbf{OntoNotes5} \cite{pradhan-etal-2013-towards} is a large-scale mixed domain NER dataset covering News, Blog and Dialogue. To make a fair comparison in the robustness experiments in Sec. \ref{sec:robust}, we use the datasets from \citet{lin-etal-2021-rockner}, which only add adversarial attack to the 11 entity types, while leaving out 7 numerical types.
    \item \textbf{WNUT17}~\cite{derczynski-etal-2017-results} is a benchmark NER dataset in social media domain. For fair comparison, we follow the data pre-processing protocols in \citet{nie-etal-2020-named}. 
    \item \textbf{Movie}~\cite{liu2013query} is a movie domain dataset containing movie queries, where long spans are annotated such as a movie’s origin or plot. We use the defaulted data split strategy into train, test sets.
    \item \textbf{Restaurant}~\cite{liu2013asgard} contains queries in restaurant domain. Similar to Movie, we use the defaulted data split strategy.
    \item \textbf{Weibo}~\cite{peng-dredze-2015-named} is a Chinese benchmark NER dataset in social media domain. We exactly follow the official data split strategy into train, dev and test sets.
\end{itemize} 

\paragraph{ABSA:} It is a fine-grained sentiment analysis task centering on aspect terms~\cite{zhang2022survey}. We explore two ABSA sub-tasks: 
\begin{itemize}[leftmargin=3mm]
    \item \textbf{Aspect Term Extraction (\textbf{ATE})} is to extract aspect terms, where there is only one query asking if there are any aspect terms in the input text.
    \item \textbf{Unified Aspect Based Sentiment Analysis} \textbf{(UABSA)} is to jointly extract aspect terms and predict their sentiment polarities. We formulate it as a SpanID task by treating the sentiment polarities, namely, positive, negative, and neutral, as three category labels, and aspect terms as spans.
\end{itemize} 
We evaluate the two sub-tasks on two datasets, including the laptop domain dataset \textbf{Lap14} and restaurant domain dataset \textbf{Rest14} from SemEval Shared tasks~\cite{pontiki-etal-2014-semeval}. We use the processed data from \citet{zhang-etal-2021-towards-generative}.
\paragraph{CCE:} It is a legal NLP task to detect and classify contract clauses into relevant clause types, such as "Governing Law" and "Uncapped Liability". The goal of CCE is to reduce the labor of legal professionals in reviewing contracts of dozens or hundreds of pages long. CCE is also a kind of SpanID task where spans are those contract clauses that warrant review or analysis and labels are predefined clause types. We conduct experiments on CCE using \textbf{CUAD}~\cite{hendrycks2021cuad}, where they annotate contracts from Electronic Data Gathering, Analysis and Retrieval (EDGAR) with 41 clause types. We follow ~\citet{hendrycks2021cuad} to split the contracts into segments within the length limitation of pretrained language models and treat each individual segment as one example. We also follow their data split strategy.

\paragraph{SBPD:} It is a typical SpanID task that aims to 
detect both the text fragment (i.e. spans) where a persuasion technique is being used as well as its technique type (i.e. category labels). We use the  \textbf{News20} and \textbf{Social21} from two SemEval shared tasks~\cite{da-san-martino-etal-2020-semeval,dimitrov-etal-2021-semeval} and follow the official data split strategy. Note that \textbf{News20} does not provide the golden label for the test set. Therefore, we evaluate \textbf{News20} on the dev set.

\subsection{Implementations}
\label{sec:impl}
We use Huggingface's implementations of BERT and RoBERTa~\cite{wolf-etal-2020-transformers}~\footnote{Chinese RoBERTa is from https://github.com/ymcui/ Chinese-BERT-wwm.}. 
The hyper-parameters can be found in Table~\ref{tab:params}.
We use Tesla V100 GPU cards for conducting all the experiments.
We follow the default learning rate schedule and dropout settings used in BERT. We use AdamW~\cite{loshchilov2018decoupled} as our optimizer.
The margin term $M$ is set to 0 for NER and ABSA, and 1 for SBPD.
The balance rate $\alpha$ is set to  $0.1$.

\begin{table}[t]
    \centering
    \small
    \setlength{\tabcolsep}{1mm}{
    \begin{tabular}{@{}lc|ccc||c|c@{}}\toprule
      \multirow{2}{*}{Methods}   & \multicolumn{4}{c||}{ \textbf{OntoNotes5}} & \multicolumn{2}{c}{ \textbf{Lap14}} \\
      \cmidrule(lr){2-5}\cmidrule(lr){6-7}
    & Ori     & \multicolumn{3}{c||}{Adv.} & Ori.    & Adv.\\
    &         & full & entity & context & & \\
    \midrule
         Tagging        & 89.8 & \textbf{56.6} & \textbf{61.9} & 83.6 & 62.3 & 44.5 \\ 
         MRC            & 90.0 & 55.3 & 61.3 & 83.3 & 63.2 & 46.9 \\
         PeerDA        & \textbf{90.1} & 55.9 & 61.0 & \textbf{84.1} & \textbf{65.9} & \textbf{50.1} \\  \bottomrule
    \end{tabular}}
    \caption{Robustness experiments against adversarial attacks. The results are reported on both original (Ori.) sets and the adversarial (Adv.) sets.}
    \label{tab:robustness}
    \vspace{-8pt}
\end{table}

\begin{figure*}
    \centering
    \includegraphics[scale=0.24]{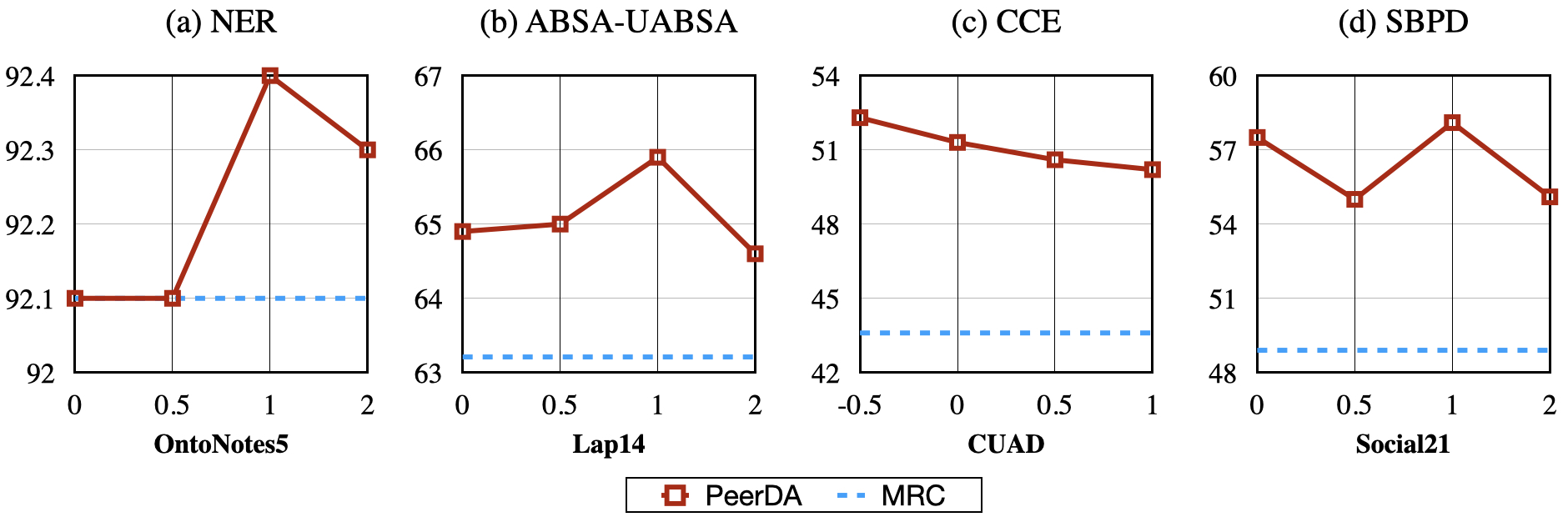}
    \caption{Performance in terms of different DA rates $\lambda$. We vary $\lambda$ to get different volumes of \textsc{PR}-based training data.} 
    \label{fig:lambda}
\end{figure*}

\subsection{Robustness}
\label{sec:robust}
To verify the advantage of PeerDA against the adversarial attack, we conduct robustness experiments using the adversarial dev set of  \textbf{OntoNotes5} \cite{lin-etal-2021-rockner} on NER and adversarial test set of \textbf{Lap14} \cite{xing-etal-2020-tasty} on UABSA.
Table \ref{tab:robustness} shows the performance on the original and the adversarial sets.
On \textbf{OntoNotes5} \textit{full} adversarial set, PeerDA improves the robustness of the model compared to MRC but slightly degrades compared to Tagging.
To investigate why this happens, we evaluate each type of adversarial attack independently, including \textit{entity} attack that replaces entities to other entities not presented in the training set and \textit{context} attack that replaces the context of entities.
It shows that PeerDA does not work well on \textit{entity} attack because we only use entities in the training set to conduct data augmentation, which is intrinsically ineffective to this adversarial attack.
This motivates us to engage outer source knowledge (e.g. Wikipedia) into our PeerDA approach in future work.
On \textbf{Lap14}, PeerDA significantly improves Tagging and MRC by 5.6 and 3.2 F$_1$ on the adversarial set respectively.

\begin{table}[t]
    \centering
    \small
    \setlength{\tabcolsep}{0.7mm}{
    \begin{tabular}{@{}lcccc@{}}
    \toprule
        Methods  & \textbf{OntoNotes5} & \textbf{Lap14} & \textbf{CUAD} & \textbf{Social21} \\ \midrule
        MRC+MenReplace  & 91.1 & 63.7 & 45.2 & 50.8 \\
        PeerDA          & \textbf{92.4} & \textbf{65.9} & \textbf{52.3} & \textbf{58.1} \\\bottomrule
    \end{tabular}}
    \caption{Performance on peer-driven DA approaches.}
    \label{tab:peer}
    \vspace{-8pt}
\end{table}

\vspace{4pt}
\subsection{Peer-driven DA}
We compare PeerDA with Mention Replacement (MenReplace)~\citep{dai-adel-2020-analysis}, another Peer-driven DA approach randomly replaces a span mention in the context with another mention of the same category in the training set. The results of four SpanID tasks are presented in Table \ref{tab:peer}. 
PeerDA exhibits better performance than MenReplace on all four tasks.
In addition, MenReplace would easily break the text coherence as a result of putting span mentions into the incompatible context, while PeerDA can do a more natural augmentation without harming the context.

\subsection{Effect of DA Rate}
\label{sec_rate}
We vary the DA rate $\lambda$ to investigate how the volume of \textsc{PR}-based training data affect the SpanID models performance.

Figure~\ref{fig:lambda} shows the effect of different $\lambda$ in four SpanID tasks. PeerDA mostly improves the MRC in all different trials of $\lambda$ and we suggest that some parameter tuning for $\lambda$ is beneficial to obtain optimal results.

Another observation is that too large $\lambda$ would do harm to the performance. Especially on CCE, due to the skewed distribution and a large number of categories, PeerDA can produce a huge size of \textsc{PR}-based training data. We speculate that too much \textsc{PR}-based training data would affect the learning of \textsc{Bl}-based training data and thus affect the model's ability to solve a SpanID task, causing the optimal $\lambda$ to be a negative value.
In addition, too much  \textsc{PR}-based training data would also increase the training cost.
As a result, we should maintain an appropriate ratio of  \textsc{Bl}-based and  \textsc{PR}-based training data to keep a reasonable performance on SpanID tasks.

\end{document}